\documentclass[runningheads]{llncs}

\usepackage{eccv}

\usepackage{eccvabbrv}

\usepackage[table]{xcolor}  

\definecolor{realblue}{RGB}{220,235,250}
\definecolor{bestred}{RGB}{255,210,210}
\definecolor{secondred}{RGB}{255,235,235}

\newcommand{\realc}[1]{\cellcolor{realblue}\textbf{#1}}
\newcommand{\bestc}[1]{\cellcolor{bestred}\textbf{#1}}
\newcommand{\secondc}[1]{\cellcolor{secondred}\textbf{#1}}

\usepackage{graphicx}
\usepackage{booktabs}
\usepackage{capt-of}
\usepackage{multirow}
\usepackage{array}
\usepackage{adjustbox}
\usepackage{amssymb}
\usepackage{amsmath}
\usepackage{pifont}    

\newcommand{\cmark}{\ding{51}} 
\newcommand{\xmark}{\ding{55}} 

\newcommand{\fitwidth}[2]{%
  \begingroup
  \setbox0=\hbox{#2}%
  \ifdim\wd0>#1
    \resizebox{#1}{!}{#2}%
  \else
    #2%
  \fi
  \endgroup
}
\usepackage[accsupp]{axessibility}  

\usepackage{hyperref}

\usepackage{orcidlink}

\begin{document}

\title{Spectral Consistent Flow for One-step 3D Medical Image Translation} 

\titlerunning{Spectral Consistent Flow}

\author{%
Haoqing Li\inst{1} \orcidlink{0000-0001-9126-5079} \and
Jun Shi\inst{1}\thanks{Corresponding author.} \orcidlink{0000-0002-9888-6238} \and
Mingchao Li\inst{1} \and
Zehua Zhu\inst{2,3} \and
Qiwei Jia\inst{1} \and
Jiong Shi\inst{4,3} \and
Hong An\inst{1}
}

\authorrunning{H. Li et al.}

\institute{%
University of Science and Technology of China, Hefei, China\\
\email{li\_haoqing@mail.ustc.edu.cn, shijun18@ustc.edu.cn}
\and
Department of Nuclear Medicine, The First Affiliated Hospital of USTC, Hefei, China
\and
Division of Life Sciences and Medicine, USTC, Hefei, China
\and
Department of Neurology, The First Affiliated Hospital of USTC, Hefei, China
}

\maketitle
\begin{abstract}
We present Spectral Consistent Flow (SC-Flow), a 3D medical image translation framework with a single function evaluation (1-NFE) in the latent space. This approach reformulates medical image translation as a stochastic Brownian bridge process that directly constructs a mapping between source and target modalities by predicting the support regularized mean velocity field. To mitigate modality entanglement, over-smoothing, and artifacts induced by the implicit low-pass modulation of the latent average velocity, we introduce a Spectral Consistency Corrector that dynamically regularizes the evolution of the power spectral density via learnable frequency-domain gain modulation. This mechanism establishes an explicit bridge between spatial textures and spectral energy flow, enabling the model to recover fine-grained anatomical fidelity while maintaining global structural coherence. Extensive experiments on four datasets demonstrate that SC-Flow delivers significantly more accurate, consistent, and robust performance across various translation scenarios.
\end{abstract}
    
\section{Introduction}
\label{sec:intro}

Medical Image Translation (MIT) aims to learn a mapping from a source modality to a target modality, enabling reduced patient scanning time and radiation exposure while supporting comprehensive multimodal diagnosis \cite{chen2025medical, xing2024cross, yang2021unified, dayarathna2024deep}.
Unlike generic unsupervised image generation \cite{goodfellow2014generative, ho2020denoising, flowmatching, ozbey2023unsupervised}, MIT benefits from paired images with precise anatomical alignment. An effective framework should leverage this pairing to model complex cross-modal relationships while maintaining anatomical fidelity. However, existing methods \cite{causalpets, ozbey2023unsupervised, pasta} struggle with this challenge and rely on strong modality- or disease-specific priors, hindering generalization across domains. Moreover, diffusion- and regression-based frameworks \cite{pasta, causalpets} are computationally expensive due to large 3D volumes and multi-step voxel-wise processing, limiting the feasibility in real-time clinical scenarios.

Flow Matching (FM) has recently emerged as a powerful alternative for learning Continuous Normalizing Flows (CNFs) by directly aligning mappings between data and prior distributions through Ordinary Differential Equations (ODEs) \cite{liu2023flow, reflow, chen2018neural, grathwohl2018ffjord, dao2023flow, albergo2023stochastic}. Its deterministic formulation allows efficient inference with only a few Neural Function Evaluations (NFEs). For instance, MeanFlow \cite{meanflow} models average velocity fields instead of marginal velocity, achieving high-quality synthesis with 1-NFE.
Despite this efficiency, applying FM to MIT remains non-trivial due to three major challenges: (1) underutilization of paired image information, where conventional FM does not explicitly integrate structural cues from the source modality, limiting fine-grained anatomical correspondence;
(2) oversmoothing and artifact formation, which degrade texture fidelity in generated images; and
(3) modality entanglement, where features from source and target domains mix, compromising anatomical or metabolic accuracy.
 
To overcome these limitations, we propose Spectral Consistent Flow (SC-Flow), an efficient 1-NFE framework for 3D medical image translation. It introduces a stochastic Brownian Bridge process \cite{brownianbridge, bbdm} to construct MeanFlow mappings across modalities, fully exploiting paired image information without relying on any modality- or disease-specific priors.
The diffusion term in the Brownian bridge injects Gaussian perturbations into the average velocity field, inducing distributional support expansion that enhances overlap across sparse samples \cite{ash2000probability, albergo2023stochastic, tong2024simulation, song2019generative}. This stochastic smoothing facilitates more stable and generalizable cross-modal mappings, particularly when paired training data is limited.
 
However, modeling the average velocity field in flow-based frameworks is not naturally compatible with FM–based MIT. The average velocity implicitly imposes a time-dependent low-pass modulation on frequency components, causing exponential attenuation of high-frequency energy \cite{hsu2011signals}. Furthermore, the high-dimensional nature of 3D medical volumes necessitates performing flow matching in latent space. The inherently low-frequency-prior characteristic of latent representations leads to systematic suppression of structural details during reconstruction \cite{ldm, park2023understanding}. Although modality bridging improves stability, the bridge flow \cite{lbm, meanflow} fails to converge precisely to the desired endpoint, retaining features from both modalities. Stochastic bridging easily induces feature entanglement between source and target domains without explicit target-modality consistency constraints, thereby compromising the anatomical and metabolic fidelity crucial for clinical interpretation.
 
According to Parseval’s theorem, the preservation of spatial textures is intrinsically linked to maintaining the energy distribution in the frequency domain \cite{brigham1988fast, hsu2011signals}. However, during flow-based generative evolution, high-frequency energy often attenuates, leading to a decay in the Power Spectral Density (PSD) that reflects texture fidelity. To address this issue, we introduce a Spectral Consistency Correction (SCC) module that adaptively modulates the radial PSD gain to compensate for high-frequency loss. Integrated end-to-end with SC-Flow, SCC enforces spectral-domain consistency complementary to voxel-domain supervision, effectively preserving texture realism and structural integrity while maintaining efficient single-step inference.
 
To summarize, our main contributions are as follows:
\begin{itemize}
    \item We propose SC-Flow for 1-NFE 3D medical image translation in the latent space. By incorporating a stochastic Brownian bridge process into average velocity modeling, SC-Flow fully exploits paired anatomical information without relying on modality- or disease-specific priors.
    \item We propose a spectral consistency correction that adaptively modulates the radial power spectrum gain to compensate for high-frequency attenuation, ensuring dual consistency in voxel and spectral domains while retaining single-step inference efficiency.
    \item Comprehensive 3D medical image translation experiments across multiple modalities and datasets demonstrating that SC-Flow achieves superior fidelity, spectral realism, and anatomical faithfulness compared with state-of-the-art diffusion, GANs, and FM-based baselines.
\end{itemize}

\begin{figure}[t]
  \centering
   \includegraphics[width=1\linewidth]{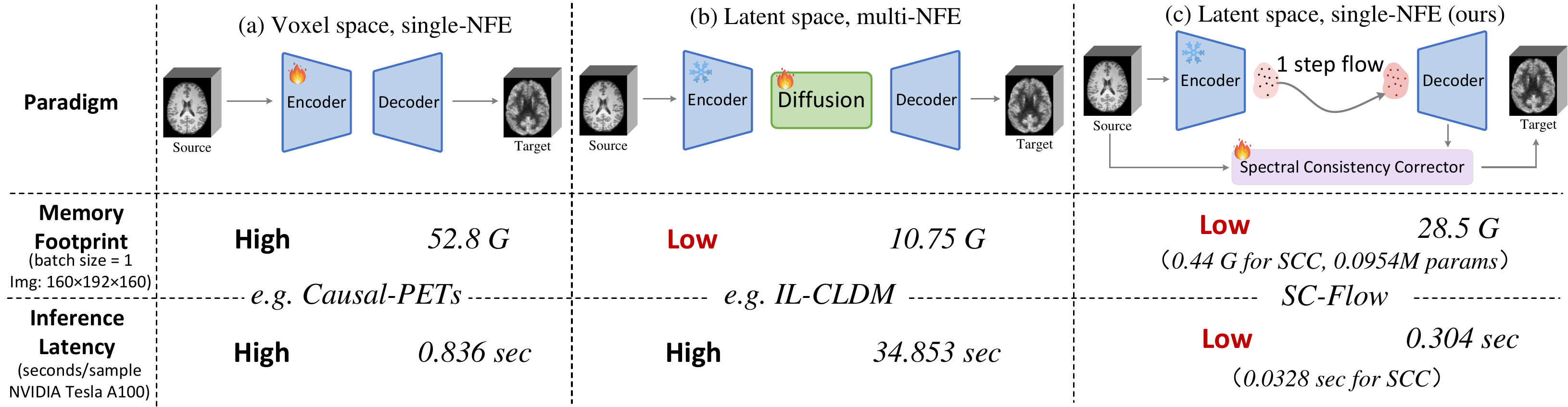}
   \caption{\textbf{SC-Flow} enables accurate, fast, and robust one-step 3D medical image translation. It achieves a lower memory footprint and higher training and inference efficiency compared with other paradigms. Benefiting from the Spectral Consistency Corrector, SC-Flow achieves latent-space 1-NFE performance that matches or even surpasses voxel-space and multi-NFE approaches.}
   \label{fig:abstract}
\end{figure}
\section{Related Work}
\label{sec:Related Work}

\textbf{Supervise and unsupervised MIT.} Recent advances have been driven by increasingly sophisticated generative frameworks.
Unsupervised methods \cite{cyclegan, park2020contrastive, wolterink2017deep, chartsias2017adversarial} establish cross-modal mappings through cycle consistency or contrastive learning, enabling image generation without precise registration. However, the absence of strict anatomical correspondence often results in geometric distortion and artifacts, limiting their clinical applicability.
Supervised methods \cite{pix2pix, cgan, causalpets, pasta, IL-CLDM} leverage paired source–target modality images acquired from the same patient, enabling voxel-wise alignment and high anatomical fidelity. \\
\textbf{2D and 3D MIT.} Despite these improvements, most methods are constrained to 2D slice-wise translation due to the prohibitive computational cost of 3D modeling \cite{wolterink2017deep, chartsias2017multimodal, dar2019image, pinaya2022brain}. However, medical images, \eg, CT, MRI, and PET, are inherently volumetric with strong anatomical continuity across adjacent slices. 2D approaches neglect these spatial dependencies, yielding inconsistent volumetric structures and degraded clinical interpretability \cite{chartsias2017multimodal, nie2017medical, yang2020unsupervised, choo2024slice}.
To address this, several 3D extensions have been proposed by expanding 2D translation networks to 3D volumes \cite{zhang2022bpgan, pasta, causalpets, lin2021bidirectional, han2023medgen3d}. Although 3D architectures improve structural continuity and volumetric realism, they dramatically increase memory usage and training time. In particular, performing 3D image translation in pixel (voxel) space incurs substantial computational and memory costs. Moreover, diffusion-based methods \cite{ho2020denoising, ldm, pasta} require multi-step noise scheduling, resulting in slow inference that hinders real-time clinical deployment.\\
\textbf{Few-NFE Diffusion and Flow Matching.} Recent progress in generative modeling has centered on reducing the heavy computational cost of diffusion and FM-based synthesis. Rather than performing hundreds of iterative denoising steps, recent methods achieve comparable fidelity with only a few NFEs. Distillation-based frameworks compress the sampling trajectory of pre-trained diffusion models into compact one/few-step mappings \cite{salimansprogressive, sauer2024adversarial, geng2023one, luo2023diff}, while Consistency Models \cite{song2023consistency, songimproved, lusimplifying, gengconsistency} eliminate teacher distillation by enforcing trajectory-level self-consistency among denoised outputs. These advances substantially accelerate inference without sacrificing image quality.

In parallel, deterministic flow-based formulations replace stochastic diffusion with continuous transport, yielding interpretable density evolution. Flow Matching \cite{flowmatching, liu2023flow, meanflow} learns time-dependent vector fields that deterministically map data to the prior via ODE integration, and MeanFlow \cite{meanflow} further an analytically solvable average velocity field, achieving single-step (1-NFE) synthesis with strong high-frequency fidelity. Despite these gains in efficiency, most existing methods overlook the anatomical correspondence intrinsic to paired medical data. Our SC-Flow explicitly incorporates cross-modal consistency within a stochastic flow formulation, ensuring structural reliability.

\section{Method}
\label{sec:method}

\begin{figure}[t]
  \centering
   \includegraphics[width=1.0\linewidth]{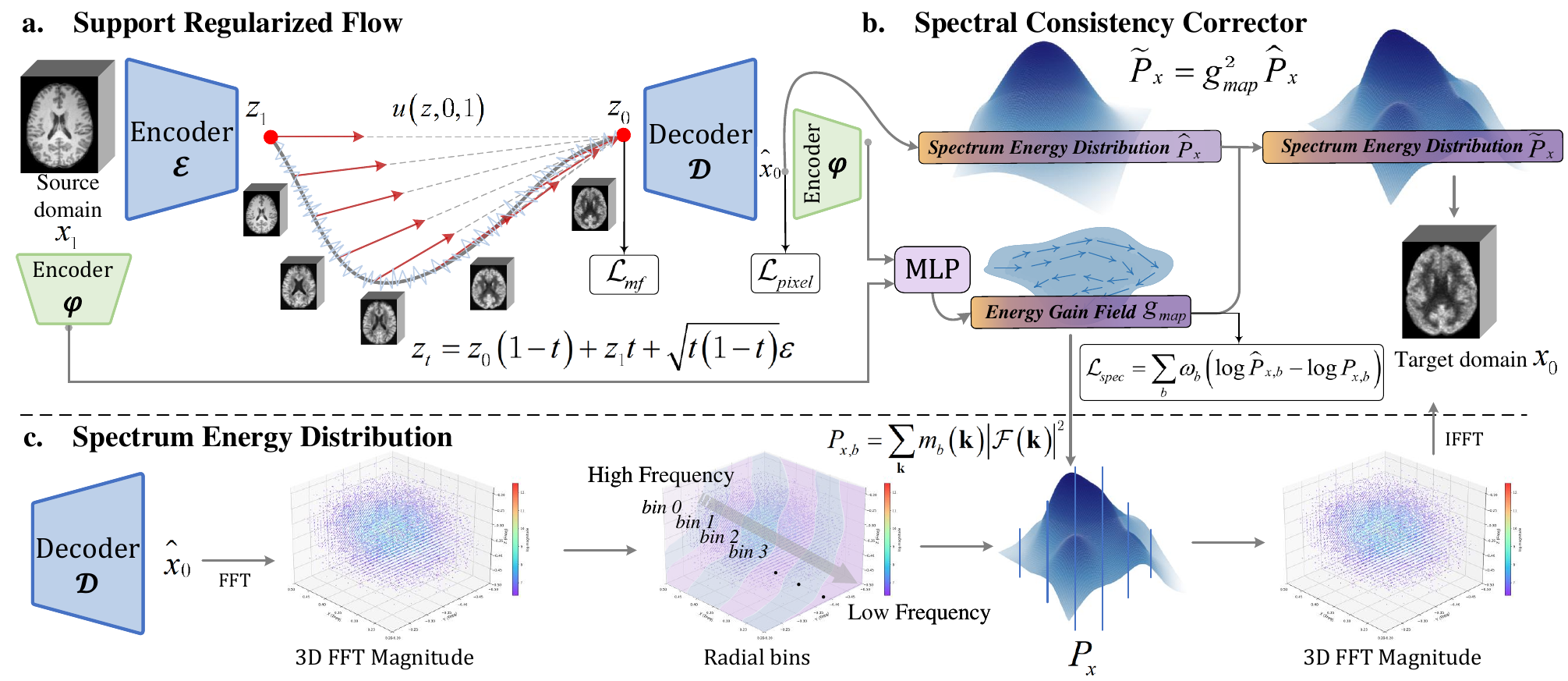}
   \caption{\textbf{Architecture Overview.} (a) Brownian Bridge Mean Flows construct a stochastic trajectory between source and target latents under the mean velocity field for one-step translation. (b) Spectral Consistency Corrector predicts adaptive frequency-domain gains to align the power spectral density and preserve texture fidelity. (c) Spectrum Energy Distribution estimates and aggregates 3D FFT magnitudes via radial binning to guide spectral correction.}
   \label{fig:main}
\end{figure}

Given two sets $X_1$ and $X_0$, sampled from distributions $p_A$ and $p_B$, respectively. The two sets form a fully matched bipartite graph, where each pair $(X_1, X_0)$ corresponds to a matched sample. The proposed Spectral Consistent Flow constructs a continuous trajectory $dz_t$ from the source domain $z_0$ to the target domain $z_1$ via a Brownian bridge process, aiming to achieve the domain transfer mapping $I_{A\xrightarrow{}B}$. This enables any $X_1$ from the domain $A$ to be transformed into its counterpart $X_0$ in the domain $B$. The overall pipeline is illustrated in Fig. \ref{fig:main}.

\subsection{Support Regularized Flow}
Flow matching methods \cite{flowmatching, liu2023flow, meanflow} typically adopt a standard Gaussian prior $p_\epsilon$ and define a reference linear interpolation path $x_t = (1-t)x + t\epsilon$ to train the velocity field mapping between data $x\thicksim p_x$ and prior $\epsilon \thicksim p_\epsilon$. However, the prior distribution is explicitly defined by paired modalities, while this unconditional FM paradigm neglects such cross-domain correspondences. In Deterministic Flow Matching (DFM), the generative process is formulated as integrating an Ordinary Differential Equation (ODE) parameterized by a neural velocity field, defining a deterministic trajectory between distributions. However, in 3D MIT tasks with limited data, the distributions $p_A$ and $p_B$ overlap only on a very narrow, high-density submanifold, leading to a severe support mismatch \cite{konz2022intrinsic}. Consequently, linear or deterministic interpolation paths are forced to traverse low-density or even void regions, resulting in unstable velocity field gradients and discontinuous distribution transitions \cite{flowmatching, albergo2023stochastic, albergobuilding, shi2023diffusion}. We extend the deterministic MeanFlow trajectory into a stochastic Brownian bridge process, which samples intermediate points under endpoint constraints, thereby regularizing the field learning and improving generalization under limited-data 3D MIT settings.
\begin{equation}
    dz_t = v_\theta(z_t, t)dt + \sigma(t)dW_t,
    \label{eq:1}
\end{equation}
where $W_t$ is a standard Wiener process defined over the time interval $[0, 1]$, and $\sigma(t)$ represents a time-dependent diffusion coefficient satisfying $\sigma(0) = \sigma(1) = 0$. The diffusion term injects Gaussian perturbations into the mean velocity field, thereby expanding the support and enabling the cross-modal trajectory to form a differentiable stochastic manifold in a statistical sense. This enhances distributional overlap while serving as an entropy regularization term in a probabilistic sense \cite{jordan1998variational, shi2023diffusion}. The marginal distribution at any intermediate time $t\in[0, 1]$ is reparameterized as:
\begin{equation}
    z_t = z_0(1-t) + z_1t+\sqrt{t(1-t)}\epsilon,
    \label{eq:2}
\end{equation}
where latent features $z=\mathcal{E}(x)$, $\mathcal{E}(\cdot)$ denotes the latent space encoder. 

As with most high-resolution image synthesis tasks \cite{ldm, jeong2025latent, lbm, bbdm}, our method operates in the latent space to reduce computational cost. The pre-trained latent space can be obtained using approaches such as Variational Autoencoder (VAE) \cite{vae} or VQGAN \cite{vqgan}, which enable sample-aware compression and decoding-based reconstruction. In the latent space, given the reparameterized intermediate bridge state $z_t$, the marginal velocity field $v_t$ can be derived as follows:
\begin{equation}
    v_t = \frac{dz_t}{dt}=(z_1-z_0)+\frac{1-2t}{2\sqrt{t(1-t)}}\epsilon,
    \label{eq:3}
\end{equation}

\begin{equation}
\begin{aligned}
     &   u(z_t,r,t) \triangleq \frac{1}{t-r}\int_{r}^{t}v(z_\tau,\tau)d\tau \\
    &= v(z_t,t)-(t-r)(v(z_t,t) \partial_z u_\theta + \partial_t u_\theta),
\end{aligned}
    \label{eq:4}
\end{equation}

\begin{equation}
    z_r = z_t-(t-r)u(z_t,r,t),
    \label{eq:5}
\end{equation}
where $v(z_\tau,\tau)$ is the marginal velocity. Average velocity $u$ serves as the underlying ground-truth field for learning and is approximated by a neural network parameterized by $\theta$ \cite{meanflow}. The predicted target domain latent representation $\hat{z}_0$ is then sampled according to Eq. \eqref{eq:5}, from which the corresponding volume $\hat{x}_0$ is reconstructed via the decoder.

\subsection{Spectral Consistency Corrector}
\label{sec:SCC}
We expect that as $t\xrightarrow{}0, z_0\thicksim p_B$. However, in practical training and sampling, the bridged flow does not always converge precisely to the desired endpoint \cite{lbm, meanflow}. The stochastic process, lacking target-modality consistency constraints, is prone to feature entanglement between the source and target modalities. Additionally, artifacts and over-smoothing may compromise the anatomical or metabolic fidelity of the generated images, potentially leading to incorrect clinical interpretations. To address these issues, we propose the Spectral Consistency Corrector (SCC) that applies a learnable gain field in the frequency domain. This mechanism explicitly aligns the power spectrum density (PSD) of the current state with the mean PSD of the target domain, while preserving low-frequency structural integrity. Consequently, the energy transfer of the Brownian bridge is closed in the frequency domain, ensuring spectral coherence throughout the process.

Specifically, let $x(\mathbf{r})$ denote a three-dimensional scalar field, where $\mathbf{r}=(x,y,z)$ represents the spatial coordinates. According to the Wiener–Khinchin theorem for wide‐sense stationary processes, under appropriate integrability conditions, the Fourier transform of the autocorrelation function equals the PSD \cite{wiener1930generalized, khintchine1934korrelationstheorie}. Accordingly, we construct the PSD function $S_x(\mathbf{k})$ for the target domain volumetric data $\hat{x}_0$:
\begin{equation}
    S_x(\mathbf{k})=|\mathcal{F}_x(\mathbf{k})|^2,
    \label{eq:6}
\end{equation}
where $\mathcal{F}_x(\mathbf{k}) = \sum_r x(\mathbf{r})e^{-2\pi i\mathbf{kr}}$ is the 3D Discrete Fourier Transform (DFT) of the volume data, computed via the Fast Fourier Transform (FFT) algorithm, and $\mathbf{k}=(k_x,k_y,k_z)$ represents the frequency vector. According to Parseval’s identity, the total energy in the spatial domain is equivalent to that in the frequency domain \cite{brigham1988fast, hsu2011signals}. Therefore, $S_x(\mathbf{k})$ naturally characterizes the energy distribution of the image across different frequency scales, enabling the assessment of deficiencies or exaggerations in high-frequency details and low-frequency structures.

\begin{figure}[t]
  \centering
   \includegraphics[width=1.0\linewidth]{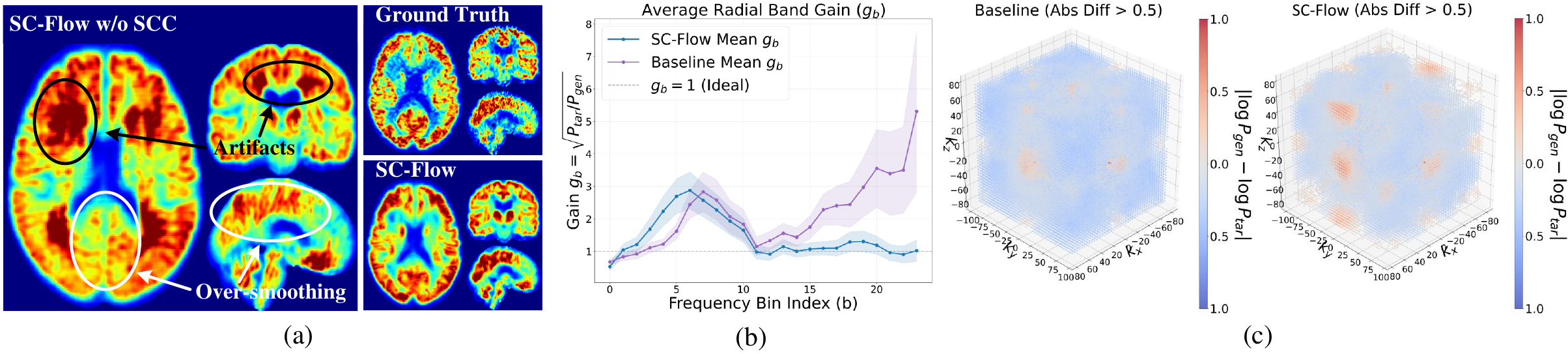}
   \caption{\textbf{Effectiveness of the Radial Gain ($g_b$) for spectral correction.} Smaller bins are low frequencies, while larger bins are high frequencies. (a) SC-Flow corrects the baseline's artifacts and high-frequency attenuation. (b) The baseline gain confirms this attenuation ($g_b \gg 1$), while SC-Flow remains near the ideal $g_b=1$. (c) 3D Frequency difference maps show the baseline's widespread spectral errors (left) are effectively corrected by our SCC, resulting in a significantly sparser points (right). The preservation of more red spots in SC-Flow proves the effectiveness of the high-frequency gain.}
   \label{fig:gain}
\end{figure}

However, due to the large variance of single-point estimates, directly computing the PSD at each frequency point $\mathbf{k}$ is both computationally expensive and unstable. Therefore, we adopt radial binning, which aggregates the 3D spectrum into several bins based on the frequency magnitude $r=|\mathbf{k}|$. Let $m_b(\mathbf{k})$ denote the soft mask of the $b$-th frequency bin, satisfying $\sum_\mathbf{k} m_b(\mathbf{k})=1$. The total energy within this bin is then defined as:
\begin{equation}
    P_{x,b} \triangleq \sum_\mathbf{k}m_b(\mathbf{k})S_x(\mathbf{k}) = \sum_\mathbf{k}m_b(\mathbf{k})|\mathcal{F}_x(\mathbf{k})|^2.
    \label{eq:7}
\end{equation}

Medical volumetric data often exhibit anisotropic sampling and direction-dependent structures. To obtain rotation-invariant spectral representations, we radially average the PSD and aggregate it across adjacent frequency bins. This preserves frequency–scale characteristics, suppresses directional noise, and reduces estimation variance, yielding a compact set of interpretable energy scalars that neural networks can learn efficiently.

To compensate for high-frequency attenuation induced by the average velocity field and to align the spectral energy distribution between the generated and target domains, we introduce a learnable gain coefficient $g_b$ for each frequency band applied to the Fourier coefficients. A continuous spectral gain map $g_{map}$ is constructed by aggregating the gains, enabling smooth frequency-dependent modulation. Since the power spectrum is proportional to the squared amplitude, the corrected spectral energy is formulated as:
\begin{equation}
    \tilde{P}_x \triangleq g_{map}^2 \hat{P}_x = (\sum_{b=1}^B g_bm_b(\mathbf{k}))^2 \hat{P}_x,
    \label{eq:8}
\end{equation}
where $\tilde{P}_x$ is the spectral energy distribution of the generator output, and $\hat{P}_x$ is the energy in the $b$-th bin of the generated image after applying the gain $g_b$. 

Our objective is to optimize $g_b$ such that the corrected $\tilde{P}_x$ approximates the ground-truth spectral energy $P_0$. The strategy for allocating gains $g_b$ across frequency bands depends on the source modality and inter-subject variability. Identical target domain spectral distortions require different high-frequency enhancement magnitudes, given differing source information or lesion presentations. Therefore, we employ a lightweight encoder $\varphi(\cdot)$ to produce a joint conditioning vector $[\varphi(\hat{x}_0),\varphi(x_1)]$. An MLP predicts the band-wise gain coefficients, yielding a frequency domain correction field $g_{map}$ and the corrected spectral energy distribution, formalized in Eq. \eqref{eq:8}. Applying the inverse Fourier transform to the corrected spectrum completes the mapping from source to target.

\subsection{Training Objective and Sampling}
\label{sec:TrainingSampling}
The training process is performed by minimizing the objective $\mathcal{L}_{mf}$ formulated as Eq. \eqref{eq:10}. In addition, to fully exploit the valuable paired medical images, we introduce the voxel loss $\mathcal{L}_{voxel}$ and spectral loss $\mathcal{L}_{spec}$ for end-to-end generation. The voxel loss is typically implemented as $L_1$ \cite{pix2pix}, perceptual loss \cite{johnson2016perceptual}, Structural Similarity (SSIM) loss \cite{wang2004image}, adversarial losses \cite{goodfellow2014generative} or other modality-specific objectives designed to capture characteristics of the medical images. 

Recently, several exploratory works \cite{leng2025repa, cohen2022diffusion} have demonstrated end-to-end latent generative training by jointly optimizing latent prior and diffusion-based generator, enabling efficient training. Inspired by these, we adopt a similar end-to-end strategy, jointly training the latent space decoder with the SC-Flow framework to refine volumetric generation in both spatial and spectral domains.

To achieve scale invariance with respect to energy ratio differences and convert multiplicative bias into additive error, we adopt the logarithmic energy difference as the core metric of the spectral objective function. Specifically, we employ a weighted log mean square error for each corrected frequency bin energy $\hat{P}_{x,b}$ and target energy $P_{x,b}$, where $w_b$ is the bin weight:
\begin{equation}
    \mathcal{L}_{spec} = \sum_b w_b(log\hat{P}_{x,b}-logP_{x,b}).
    \label{eq:9}
\end{equation}
The final training objective is summarized as follows:
\begin{equation}
    \mathcal{L}_{mf}(z_t,r,t) = \mathbb{E}||u_\theta(z_t,r,t)-u(z_t,r,t)||_2^2,
    \label{eq:10}
\end{equation}
\begin{equation}
    \mathcal{L} = \lambda_1\mathcal{L}_{mf}(z_t,r,t) + \lambda_2\mathcal{L}_{voxel}(\hat{x}_0,x
    _0) + \lambda_3\mathcal{L}_{spec}(\hat{x}_0,x_0),
    \label{eq:11}
\end{equation}
where $\lambda_1$, $\lambda_2$, and $\lambda_3$ are hyperparameters that weigh each loss component. We employ single-NFE sampling, \ie, $r=0, t=1$. In this case, both the term $z_0(1-t)$ and the noise component $\sqrt{t(1-t)}\epsilon$ vanish, allowing a simplified one-step sampling process expressed as $z_0 = z_1-u_\theta(z_1,0,1)$. 
\section{Experiments}

\subsection{Experimental Setting}

\begin{table*}[t]
    \centering
    \caption{\textbf{Quantitative comparison} of MRI (T1WI and FLAIR) to FDG, Tau, and A$\beta$-PET translation on the NACC, A4, and ADNI datasets.}
    \label{tab:comparison}
    \resizebox{\textwidth}{!}
    {
    \begin{tabular}{lcccccccccccccc}
    \toprule
    \multirow{2}{*}{Method} & 
    \multicolumn{1}{c}{} &
    \multicolumn{4}{c}{\textbf{MRI to A$\beta$-PET (ADNI)}} & 
    \multicolumn{4}{c}{\textbf{MRI to FDG-PET (NACC)}} &
    \multicolumn{4}{c}{\textbf{MRI to Tau-PET (A4)}} \\
    
    \cmidrule(lr){3-6} \cmidrule(lr){7-10} \cmidrule(lr){11-14}
     & NFE & MAE $\downarrow$ & RMSE $\downarrow$ & PSNR$\uparrow$ & SSIM $\uparrow$ & MAE $\downarrow$ & RMSE $\downarrow$ & PSNR$\uparrow$ & SSIM $\uparrow$ & MAE $\downarrow$ & RMSE $\downarrow$ & PSNR$\uparrow$ & SSIM $\uparrow$ \\
    \midrule
    \midrule
    \multicolumn{13}{c}{\textit{Multi-NFE}} \\
    \midrule
    \midrule
    LDM~\cite{ldm} & 1000 & 6.35&11.57&19.03&63.60 & 18.35&26.06&11.81&43.09 & 14.41&22.30&13.07&49.32 \\
    IL-CLDM~\cite{IL-CLDM}&1000&4.13&8.34&22.46&77.34 & 5.10&10.44&20.04&83.93 & 5.52&	10.70&20.22&82.02&\\
    PASTA~\cite{pasta} & 100 & 3.63&7.61&23.04&85.18 &3.38&7.75&22.43&84.64&2.84&5.90&24.99&88.48 \\
    LBM~\cite{lbm} & 100 & 3.35&7.24&23.42&85.19 & 2.59&6.26&24.17&84.77 & 3.14&6.92&23.58&83.99 \\
    CoCoLIT~\cite{Cocolit}&5& 4.83&8.03&22.71&84.29& 2.58&6.02&24.69&88.52 & 3.32&6.86&23.83&87.63\\

    \midrule
    \midrule
    \multicolumn{13}{c}{\textit{Single-NFE}} \\
    \midrule
    \midrule
    CycleGAN~\cite{cyclegan} & \bestc{1} & \secondc{3.21}&\secondc{6.91}&\secondc{23.69}&86.18 & 2.17&5.24&25.70&90.10 & 2.73&5.55&25.51&86.50 \\
    Pix2Pix~\cite{pix2pix} & \bestc{1} & 3.37&7.45&22.97&85.58 & \secondc{2.01}&4.89&26.26&91.58 & 2.77&5.98&24.67&84.90 \\
    MeanFlow~\cite{meanflow} & \bestc{1} & 3.74&8.05&22.16&78.32 & 3.38&7.99&21.96&78.18 & 3.27&7.32&22.81&79.67 \\
    Causal-PETs~\cite{causalpets} & \bestc{1} & 3.97&7.13&23.57&\secondc{87.08} & 2.03&\secondc{4.66}&\secondc{26.70}&\secondc{92.47} & \secondc{2.63}&\secondc{5.40}&\secondc{25.83}&\secondc{89.32} \\
    
    \midrule
    \midrule
    \textbf{SC-Flow (Ours)} & \bestc{1} & \bestc{2.98} & \bestc{6.32} & \bestc{24.72} & \bestc{88.71} & \bestc{1.77} & \bestc{4.34} & \bestc{27.41} & \bestc{93.30} & \bestc{2.38} & \bestc{5.09} & \bestc{26.43} & \bestc{90.33} \\
    \bottomrule
    \end{tabular}
    }
\end{table*}

\begin{figure}[t]
  \centering
   \includegraphics[width=1.0\linewidth]{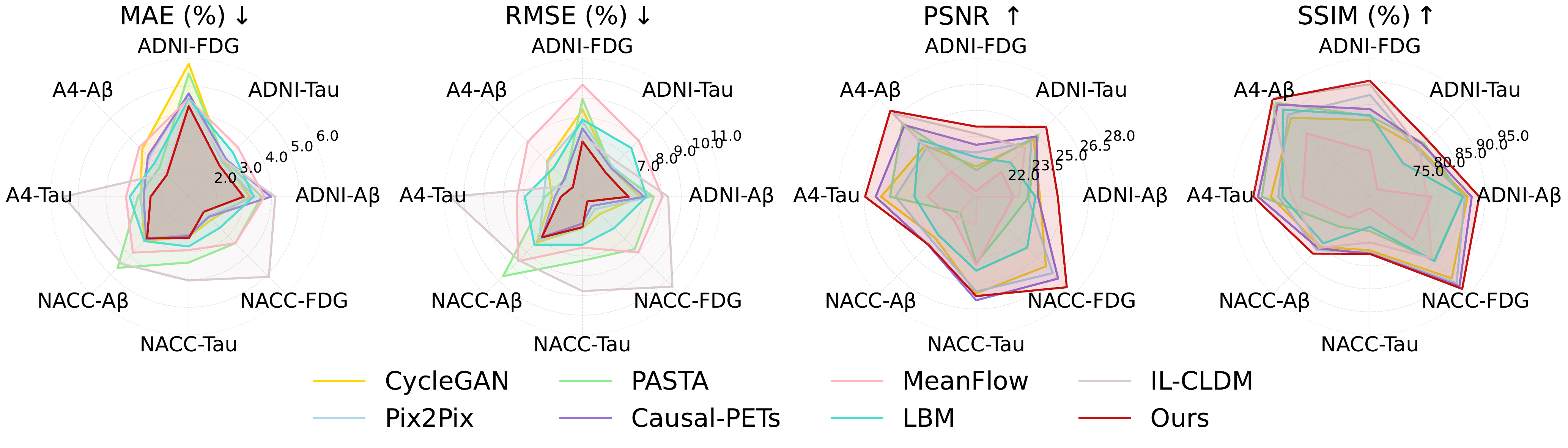}
   \caption{Comparison of all the MRI (T1WI and FLAIR) to FDG, Tau, and A$\beta$-PET scenarios on the NACC, A4 and ADNI datasets.}
   \label{fig:Radar}
\end{figure}

\begin{figure}[t]
  \centering

  \begin{minipage}[t]{0.49\textwidth}
  \vspace{0pt}
    \centering
    \includegraphics[width=1.0\linewidth]{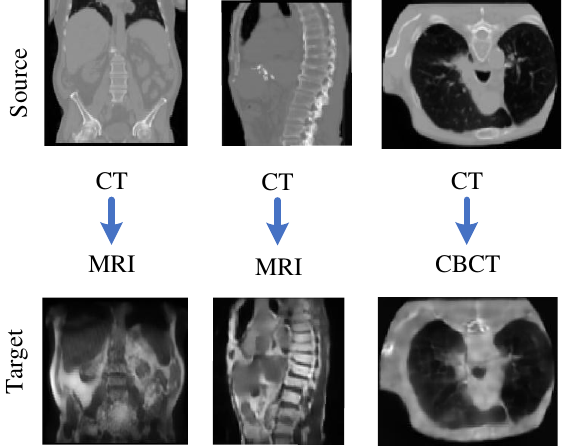}
    \captionof{figure}{CT to MRI / CBCT translation scenarios on the synthRAD2025 dataset.}
    \label{fig:show2}
  \end{minipage}
  \hfill
  \begin{minipage}[t]{0.49\textwidth}
  \vspace{0pt}
    \centering
    \captionof{table}{\textbf{Quantitative comparison} of CT to CBCT / MRI translation on the synthRAD2025 dataset.}
    \label{tab:comparison2}

    \footnotesize
    \setlength{\tabcolsep}{3.5pt}
    \renewcommand{\arraystretch}{1.05}

    \resizebox{\linewidth}{!}{
      \begin{tabular}{lcccc}
        \toprule
        \multirow{2}{*}{Method} &
        MAE $\downarrow$ & RMSE$\downarrow$ & PSNR$\uparrow$ & SSIM $\uparrow$ \\
        \cmidrule(lr){2-5}
        & \multicolumn{4}{c}{\textit{CT to MRI}} \\
        \midrule
        \midrule
        CycleGAN~\cite{cyclegan} & 7.02 & 12.24 & 19.00 & 70.46 \\
        MeanFlow~\cite{meanflow} & 4.29 & 9.76  & 20.71 & 72.20 \\
        LBM~\cite{lbm}           & \secondc{4.08} & \secondc{9.40}  & \secondc{21.05} & \secondc{74.59} \\
        \textbf{Ours}            & \bestc{3.77} & \bestc{8.53} & \bestc{21.84} & \bestc{78.40} \\
        \midrule
        & \multicolumn{4}{c}{\textit{CT to CBCT}} \\
        \midrule
        \midrule
        CycleGAN~\cite{cyclegan} & 4.41 & 7.69 & 22.85 & 82.39 \\
        MeanFlow~\cite{meanflow} & 3.62 & 6.71 & 23.99 & 82.97 \\
        LBM~\cite{lbm}           & \secondc{3.01} & \secondc{5.63} & \secondc{25.91} & \bestc{88.37} \\
        \textbf{Ours}            & \bestc{2.92} & \bestc{5.53} & \bestc{26.10} & \secondc{87.69} \\
        \bottomrule
      \end{tabular}
    }
  \end{minipage}
\end{figure}

\begin{figure}[t]
  \centering
  \begin{minipage}[t]{0.49\textwidth}
  \vspace{0pt}
    \centering
    \includegraphics[width=1.0\linewidth]{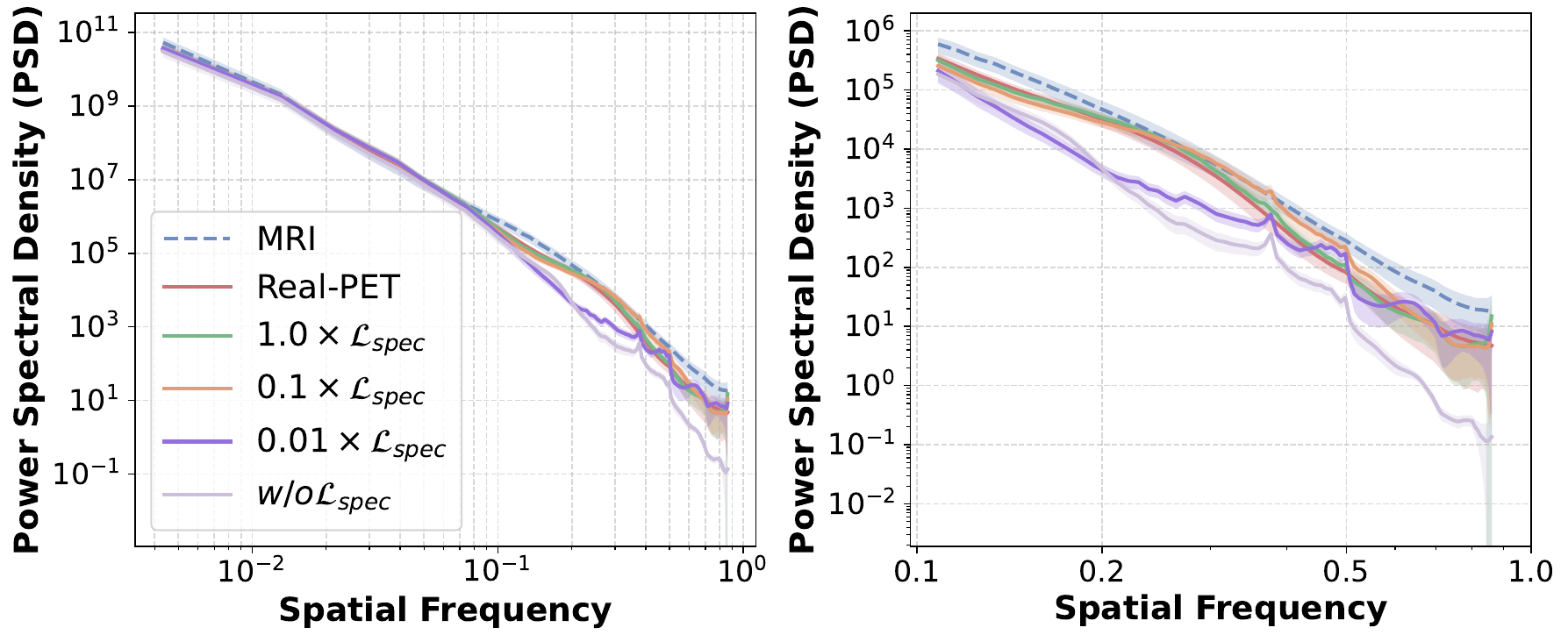}
   \caption{\textbf{Effect of the $L_{spec}$ weight on the PSD.} Without $L_{spec}$, PSD rapidly decays to 1 / 100 of the ground truth. Increasing $L_{spec}$ weight restores high-frequency energy.}
   \label{fig:PSD_lossweight}
  \end{minipage}
  \hfill
  \begin{minipage}[t]{0.49\textwidth}
  \vspace{0pt}
    \centering
    \includegraphics[width=1.0\linewidth]{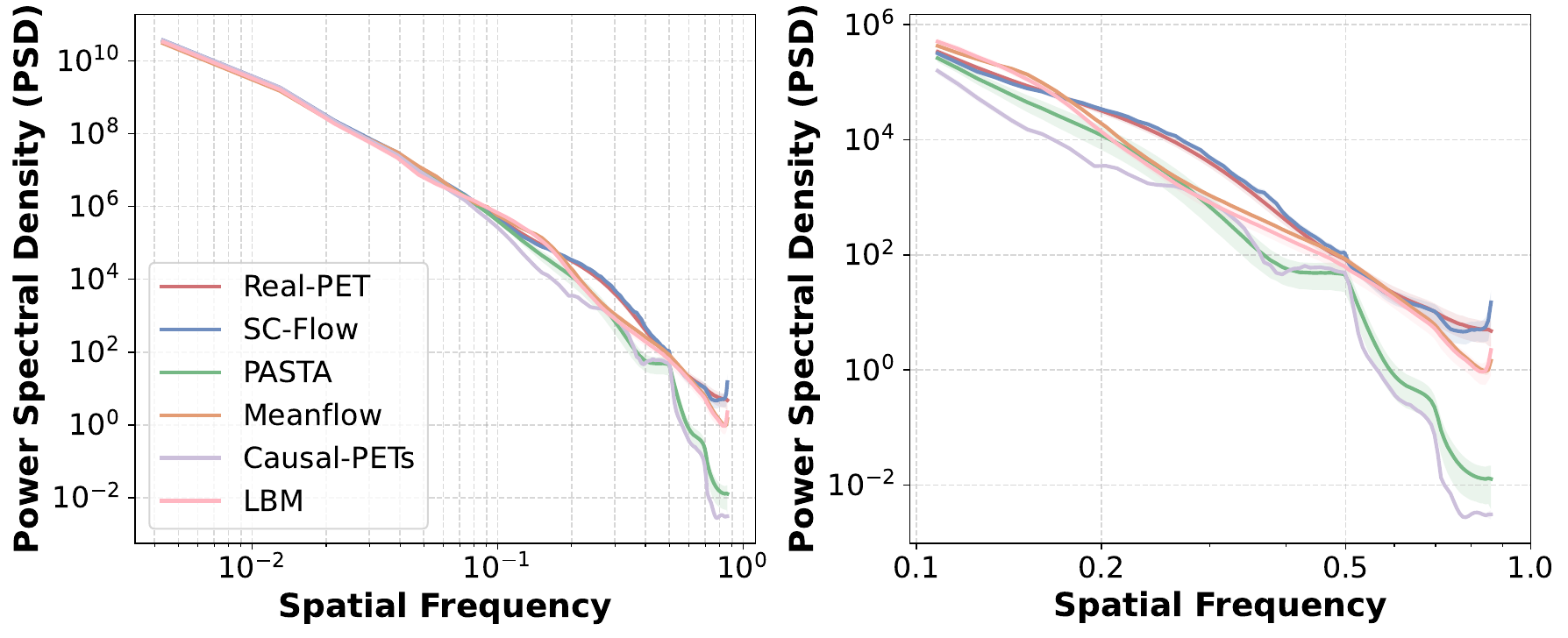}
   \caption{\textbf{Comparison of PSD curves} across different methods. Our approach is closer to the ground truth. X-axis represents frequency from low to high values. Right: zoomed-in high-frequency region.}
   \label{fig:PSD_comparison}
  \end{minipage}
\end{figure}

We evaluate our SC-Flow on eight 3D medical image translation tasks: (1) MRI to FDG-PET, (2) MRI to Tau-PET, (3) MRI to A$\beta$-PET, (4) CT to MRI, and (5) CT to CBCT, (6) A$\beta$-PET to MRI, (7) Tau-PET to  MRI, (8) FDG-PET to MRI, across four datasets to demonstrate its effectiveness. These datasets include ADNI \cite{adni}, A4 \cite{a4}, NACC \cite{nacc}, and synthRAD2025 \cite{synthrad2025}, covering 8 imaging modalities. Detailed information is provided in the supplementary materials.

We use Peak Signal-to-Noise Ratio (PSNR), Structural Similarity Index Metric (SSIM) (\%), Mean Absolute Error (MAE) (\%), Root Mean Square Error (RMSE) (\%), Fréchet Inception Distance (FID), and Maximum Mean Discrepancy (MMD) as evaluation metrics. We compare our SC-Flow to the state-of-the-art methods , including general methods CycleGAN \cite{cyclegan}, LDM \cite{ldm}, Pix2Pix \cite{pix2pix}, LBM \cite{lbm}, MeanFlow \cite{meanflow} , and medical-specific methods Causal-PETs \cite{causalpets}, PASTA \cite{pasta}, IL-CLDM \cite{IL-CLDM}, CoCoLIT \cite{Cocolit}.

Other details of the implementation and data processing are shown in the Supplementary Material.

\subsection{Quantitative Results}

\begin{figure}[t]
  \centering
  \includegraphics[width=1.0\linewidth]{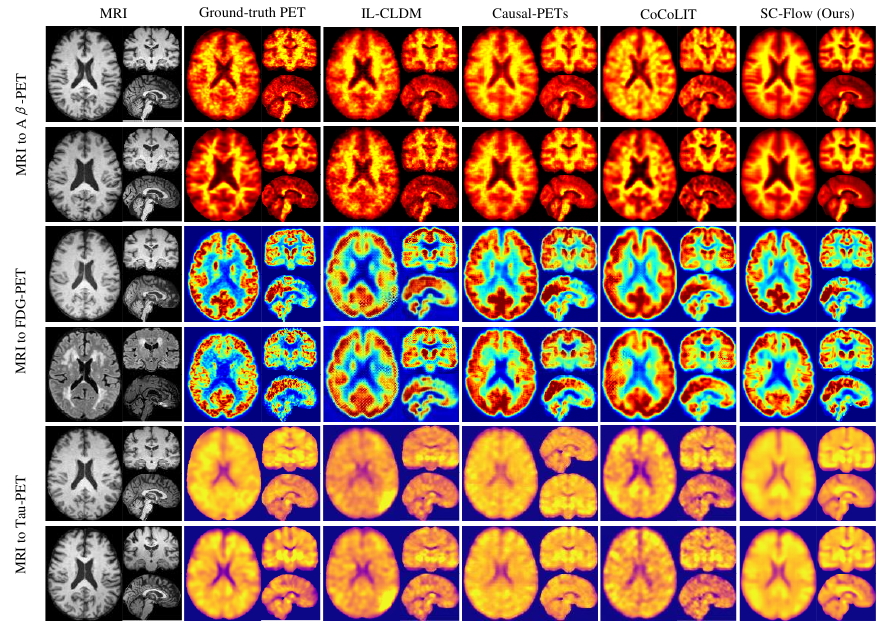}
  \caption{\textbf{Qualitative comparison} on MRI (T1WI and FLAIR) to FDG, Tau, and A$\beta$-PET translation scenarios on three datasets.}
  \label{fig:show1}
\end{figure}

We first evaluate SC-Flow across the NACC, A4, and ADNI datasets for MRI to FDG, Tau, and A$\beta$-PET translation. The results are shown in Table \ref{tab:comparison} and Fig. \ref{fig:Radar}. Our approach achieves the best medical image translation accuracy in all translation scenarios, while being competitive in terms of running time. SC-Flow consistently outperforms all multi- and single-NFE baselines across every dataset and modality. Compared to the previous MeanFlow model, SC-Flow reduces MAE by more than 20\% and improves PSNR by more than 2.5 dB, indicating superior voxel-wise reconstruction accuracy and enhanced signal fidelity. In the supervised setting, our approach surpasses medical-specific models such as Causal-PETs and PASTA, which demonstrate better preservation of fine anatomical and metabolic structures. Notably, SC-Flow demonstrates strong robustness across multiple datasets. We further evaluate it on the NACC dataset for FDG, Tau, and A$\beta$-PET to MRI translation, where it consistently delivers stable performance across all three translation settings (Table \ref{tab:trans_nacc}). Comprehensive results across diverse modalities and datasets are in the Supplementary Material.

Fig. \ref{fig:PSD_comparison} compares the power spectral density (PSD) of reconstructed PET images across different methods. Baseline approaches such as Causal-PETs and PASTA exhibit steep high-frequency decay, leading to artifacted and over-smoothed reconstructions. FM-based methods \cite{lbm, meanflow} alleviate this issue to some extent but still diverge from the real PET spectrum, particularly in the mid-to-high-frequency range. In contrast, SC-Flow achieves the closest PSD alignment with the ground truth across all frequencies, indicating superior preservation of both low-frequency structural information and high-frequency texture fidelity. These results validate that SCC effectively stabilizes spectral energy evolution during the flow, yielding physically consistent and perceptually realistic reconstructions.

We conducted comparative experiments of CT to MRI and CBCT translation on the synthRAD2025 dataset to validate the effectiveness of our method across more modalities and anatomical regions, as shown in Table \ref{tab:comparison2}. Our approach achieves a superior balance between global intensity accuracy and structural fidelity, evidenced by simultaneously achieving significantly better performance across all scenarios. This robust improvement, particularly pronounced on the MRI task, is directly attributed to the spectral consistency constraint.

\textbf{Downstream diagnostic tasks.} Tab.~\ref{tab:downstream_tasks} evaluates MIT quality by training classifiers on real images and testing them on images generated by each method for Alzheimer's Disease versus Cognitively Normal diagnosis. This setting measures both practical diagnostic utility and the synthetic-to-real gap. Across ResNet50 \cite{resnet} on NACC-FDG and SeResNext \cite{SeResNext} on A4-Tau, SC-Flow achieves the best diagnostic performance. It improves F1 by +11.45 over the strongest baseline on NACC-FDG and AUC by +24.04 on A4-Tau, while also obtaining the best or near-best fidelity and divergence scores. These results show that SC-Flow's improved perceptual alignment translates into stronger downstream clinical utility.

\begin{table}[t]
    \centering
    \caption{Downstream diagnostic tasks.}
    \label{tab:downstream_tasks}
    
    \fontsize{7}{8}\selectfont
    \setlength{\tabcolsep}{3.2pt}
    \renewcommand{\arraystretch}{1.12}
    
    \begin{adjustbox}{max width=\linewidth}
    \begin{tabular}{p{18mm}@{\hspace{6pt}}lccccccc}
    \toprule
    \multicolumn{2}{c}{\textbf{Method}} &
    \textbf{F1} & \textbf{AUC} & \textbf{Acc} & \textbf{Prec} &
    \textbf{Recall} & \textbf{FID}$\downarrow$\% & \textbf{MMD}$\downarrow$ \\
    \midrule
    \midrule
    \multirow{7}{*}{\parbox[c]{18mm}{\centering\textbf{ResNet50 \cite{resnet}}\\\textbf{(NACC-FDG)}\strut}}
    & \realc{Real}       & \realc{89.23} & \realc{78.13} & \realc{97.14} & \realc{98.48} & \realc{83.33} & \realc{--} & \realc{--} \\
    & Pix2Pix \cite{pix2pix}            & 59.77 & \bestc{82.29} & 77.14 & 59.19 & \secondc{72.40} & \bestc{0.09} & \bestc{0.1269} \\
    & CycleGAN \cite{cyclegan}           & \secondc{60.32} & 73.96 & \secondc{85.71} & \secondc{59.27} & 61.98 & 0.18 & 0.2029 \\
    & LBM \cite{lbm}                & 43.55 & 38.54 & 77.14 & 45.00 & 42.19 & 0.45 & 0.2145 \\
    & CausalPETs \cite{causalpets}         & 46.97 & 70.83 & \bestc{88.57} & 45.59 & 48.44 & 0.69 & 0.3658 \\
    & MeanFlow \cite{meanflow}          & 43.29 & 44.79 & 60.00 & 49.30 & 47.92 & 1.48 & 0.4700 \\
    & \textbf{SC-Flow (Ours)}     & \bestc{71.77} & \secondc{81.25} & \bestc{88.57} & \bestc{68.33} & \bestc{78.65} & \secondc{0.16} & \secondc{0.1384} \\
    \midrule
    \midrule
    \multirow{7}{*}{\parbox[c]{18mm}{\centering\textbf{SeResNext \cite{SeResNext}}\\\textbf{(A4-Tau)}\strut}}
    & \realc{Real}       & \realc{61.43} & \realc{56.01} & \realc{73.33} & \realc{66.73} & \realc{60.70} & \realc{--} & \realc{--} \\
    & Pix2Pix \cite{pix2pix}           & 42.97 & 38.22 & 51.11 & 43.33 & 42.79 & \secondc{0.54} & 0.3902 \\
    & CycleGAN \cite{cyclegan}          & 38.68 & 42.91 & 53.33 & 37.86 & 39.78 & 0.93 & 0.4674 \\
    & LBM \cite{lbm}               & 36.62 & \secondc{46.15} & \secondc{57.78} & 33.33 & 40.63 & 2.47 & 0.5307 \\
    & CausalPETs \cite{causalpets}        & 38.68 & 24.04 & 53.33 & 37.86 & 39.78 & 0.73 & 0.4874 \\
    & MeanFlow \cite{meanflow}          & \secondc{48.16} & 45.43 & 55.56 & \secondc{48.33} & \secondc{48.20} & 0.85 & \secondc{0.2831} \\
    & \textbf{SC-Flow (Ours)}     & \bestc{59.63} & \bestc{70.19} & \bestc{71.11} & \bestc{62.84} & \bestc{59.13} & \bestc{0.32} & \bestc{0.1176} \\
    \bottomrule
    \end{tabular}
    \end{adjustbox}
\end{table}

\begin{figure}[t]
  \centering
  \noindent
  \begin{minipage}[t]{0.62\textwidth}
  \vspace{0pt}
    \centering
    \includegraphics[width=\linewidth]{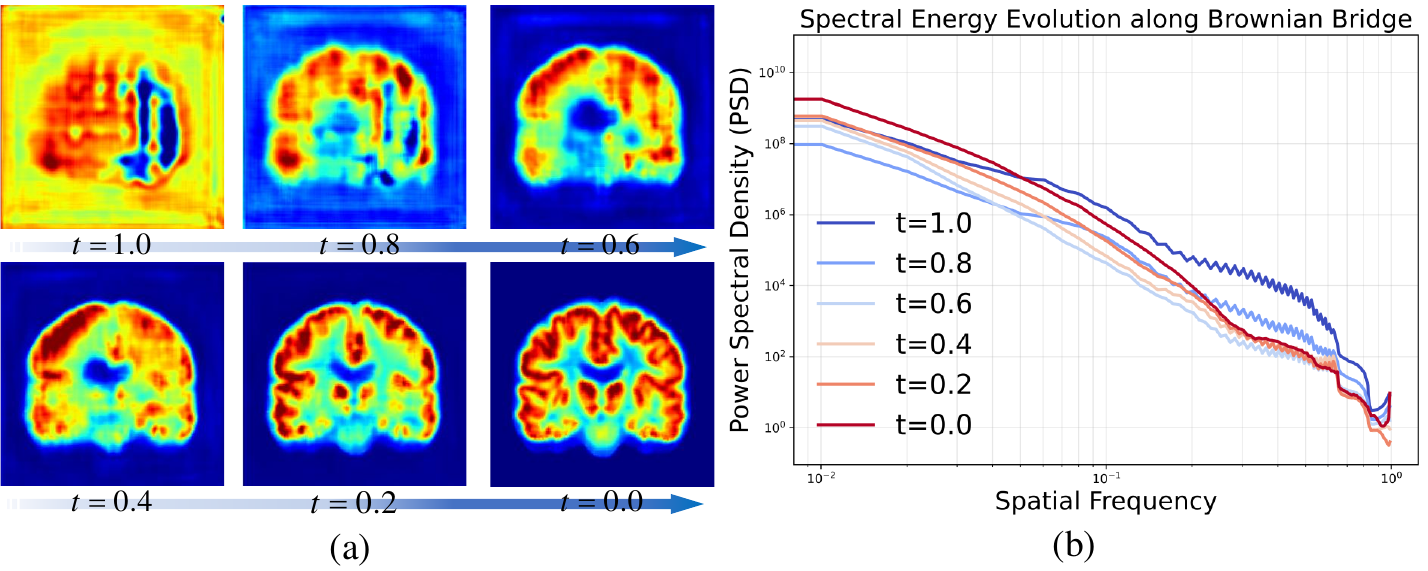}
  \end{minipage}\hfill
  \begin{minipage}[t]{0.35\textwidth}
  \vspace{0pt}
    \raggedright
    \caption{Evolution of intermediate bridge states over timesteps, illustrating how the Brownian-bridge trajectory progressively transports the source modality toward the target distribution.}
    \label{fig:t}
  \end{minipage}
\end{figure}

\subsection{Qualitative Results}

\begin{figure}[t]
  \centering
   \includegraphics[width=1.0\linewidth]{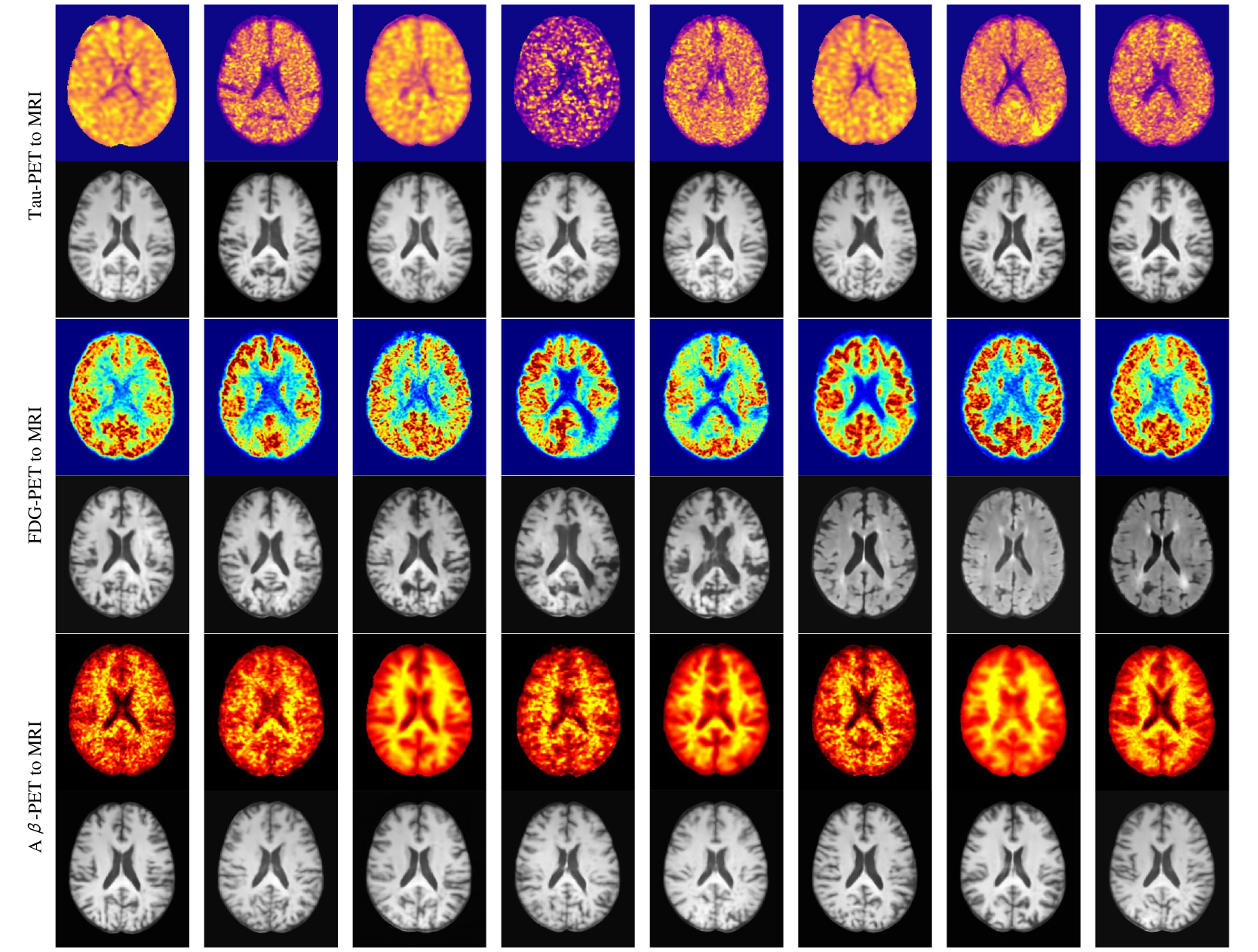}
   \caption{FDG, Tau, and A$\beta$-PET to MRI translation on the NACC dataset.}
   \label{fig:show3}
\end{figure}

Visual comparisons of 3D translation results across the four benchmarks are presented in Fig. \ref{fig:show1} and Fig. \ref{fig:show2}. Our method consistently yields results closest to the ground truth. Compared with LBM, Causal-PETs, and MeanFlow, our method generates more realistic PET appearances with clear cortical boundaries and accurate metabolic distributions. The baseline methods tend to introduce unnatural artifacts and noisy textures, while our approach effectively suppresses these degradations through the proposed frequency-consistent correction.
Specifically, for the MRI→A$\beta$-PET task, SC-Flow restores the subtle $\beta$-amyloid deposition pattern in the frontal and temporal cortices that is almost lost in MeanFlow. For the MRI→FDG-PET task, our results preserve the fine-grained metabolic contrast between gray and white matter, showing reduced halo artifacts and sharper regional metabolism compared to others. For the MRI→Tau-PET task, our method maintains a smooth yet anatomically faithful uptake distribution, while others exhibit visible intensity bias or blurred cortical rims.

Moreover, our method avoids the modality entanglement common in unsupervised flow-based models, where structural MRI features are inadvertently introduced into the PET domain. The proposed spectral constraint decouples the modality-specific frequency bands, ensuring that the generated PET images retain only target-domain features.
Across all datasets (ADNI, NACC, A4), our results demonstrate consistent cross-modality robustness, preserving the diagnostic cues essential for clinical interpretation.

Additionally, we performed reverse translation experiments on the NACC dataset, \ie, translating from smoother functional images (A$\beta$-PET, Tau-PET, and FDG-PET) to structural MRI with more tissue details. As illustrated in Fig. \ref{fig:show3}, our method demonstrates superior performance across a range of experimental settings. 
Our method obtains the high fidelity MRI, with well-defined gray matter-white matter contrast and accurate representation of subcortical structures such as the hippocampus and thalamus. Moreover, our method demonstrates robustness across different tracers and imaging conditions, as evidenced by the consistent quality of the translated MRI images regardless of the PET tracer or the intensity of the PET signal.

\subsection{Ablation Study}
We perform an ablation study to validate major design choices for our SC-Flow and spectral consistency corrector module. Specifically, we evaluate SC-Flow results with different configurations: 1) vanilla baseline, 2) SC-Flow without $\mathcal{L}$ / SCC, 3) SC-Flow without diffusion term (w/o $dW_t$), 4) without spectral consistency corrector (w/o SCC), etc., as shown in Table \ref{tab:ablation_comp}, our full system (bins = 24) outperforms all other alternative configurations.

We examine the effect of radial bin settings in SCC, as shown in Table \ref{tab:bins_nacc}. Excessively fine bins (\eg, 96) cause instability and overfitting, while too coarse discretization underrepresents spectral variation. A moderate bins (\eg, 24) achieves the best trade-off between accuracy and stability.

\begin{figure}[t]
  \centering
  \noindent
  \begin{minipage}[t]{0.49\linewidth}
  \vspace{0pt}
    \centering

    \captionof{table}{\textbf{Component ablation} on the NACC dataset, evaluating the contribution of each module ($dW_t$, SCC, and $\mathcal{L}$) to translation quality. $\checkmark/\times$ indicate whether a component is enabled. Lower MAE/RMSE and higher PSNR/SSIM denote better performance.}
    \label{tab:ablation_comp}

    \scriptsize
    \setlength{\tabcolsep}{3pt}
    \renewcommand{\arraystretch}{1.08}

    \fitwidth{\linewidth}{%
    \renewcommand{\arraystretch}{1.13}
    \begin{tabular}{
      >{\centering\arraybackslash}p{4mm}
      >{\centering\arraybackslash}p{4mm}
      >{\centering\arraybackslash}p{4mm}
      cccc
    }
      \toprule
      $dW_t$ & SCC & $\mathcal{L}$ &
      MAE & RMSE & PSNR & SSIM \\
      \midrule
      \midrule
      \xmark & \xmark & \xmark & 3.38 & 7.99 & 21.97 & 78.18 \\
      \cmark & \xmark & \xmark & 2.47 & 5.96 & 24.59 & 86.54 \\
      \xmark & \cmark & \xmark & 2.57 & 6.19 & 24.25 & 84.33 \\
      \xmark & \cmark & \cmark & 2.50 & 6.03 & 24.47 & 85.69 \\
      \cmark & \cmark & \xmark & 2.42 & 5.83 & 26.14 & 87.72 \\
      \cmark & \xmark & \cmark & 2.11 & 5.15 & 25.86 & 89.11 \\
      \cmark & \cmark & \cmark & 1.77 & 4.34 & 27.41 & 93.30 \\
      \bottomrule
    \end{tabular}%
    }
    \renewcommand{\arraystretch}{1.08}

  \end{minipage}\hspace{0.03\linewidth}
  \begin{minipage}[t]{0.48\linewidth}
  \vspace{0pt}
    \centering

    \scriptsize
    \setlength{\tabcolsep}{3pt}

    \captionof{table}{FDG, Tau and A$\beta$-PET$\to$MRI translation on NACC.}
    \label{tab:trans_nacc}

    \renewcommand{\arraystretch}{1.20}
    \begin{tabular}{lcccc}
      \toprule
      \textit{trans.} & MAE & RMSE & PSNR & SSIM \\
      \midrule
      \midrule
      A$\beta$$\to$MRI & 2.58 & 6.49 & 24.05 & 90.46 \\
      Tau$\to$MRI      & 2.73 & 6.83 & 23.56 & 88.53 \\
      FDG$\to$MRI      & 2.73 & 6.81 & 23.52 & 88.66 \\
      \bottomrule
    \end{tabular}

    \captionof{table}{Sensitivity to spectral bin number on NACC.}
    \label{tab:bins_nacc}

    \renewcommand{\arraystretch}{1.05}
    \begin{tabular}{lcccc}
      \toprule
      Bins & MAE & RMSE & PSNR & SSIM \\
      \midrule
      \midrule
      12 & 1.84 & 4.48 & 27.08 & 92.50 \\
      24 & 1.81 & 4.39 & 27.24 & 92.77 \\
      48 & 1.83 & 4.44 & 27.12 & 92.77 \\
      96 & 2.10 & 5.08 & 25.97 & 88.58 \\
      \bottomrule
    \end{tabular}

  \end{minipage}
\end{figure}
\section{Discussion and Conclusion}

\textbf{Conclusion.}
We presented SC-Flow, a 1-NFE framework for 3D medical image translation. It formulates translation as an endpoint-conditioned Brownian bridge in latent space and learns a mean-flow field for one-step generation. With a spectral consistency corrector, SC-Flow better preserves anatomical textures and reduces modality entanglement across datasets and translation settings.\\
\textbf{Limitations.}
SC-Flow relies on the autoencoder latent space, so AE information loss can limit perceptual fidelity, especially for subtle structures or low-quality scans. The spectral corrector reduces the drift but cannot fully compensate for severe modality gaps, imperfect latent representations, or limited training data. Future work will improve latent representation learning, integrate spectral priors into the flow dynamics, and extend it to spatiotemporal disease progression.

\section*{Acknowledgements}
This work was supported in part by the Xiaomi Young Scholars Program.

%
%
\bibliographystyle{splncs04}
\bibliography{main}
\end{document}